\documentclass[journal]{IEEEtran}
\usepackage{graphicx}
\usepackage{enumerate}
\usepackage{amsmath,amsfonts,amssymb}
\usepackage{color}
\usepackage{multirow}
\usepackage{float}
\usepackage[noend]{algpseudocode}
\usepackage{algorithmicx,algorithm}
\usepackage{setspace}
\usepackage{lineno,hyperref}
\hypersetup{hidelinks}
\usepackage[nameinlink,noabbrev]{cleveref}
\usepackage{booktabs}
\usepackage[table]{xcolor}
\usepackage{threeparttable}
\modulolinenumbers[5]
\usepackage{etoolbox}
\begin{document}
%\begin{frontmatter}

%倩倩todo 在下面添加标题替换
% \title{EdgeMamba-LRD: Low-Rank Structured 2D State Space Modeling with Structure-Aware Distillation for Edge-Efficient Multispectral Small-Target Detection}
\title{DLRMamba: Distilling Low-Rank Mamba for Edge Multispectral Fusion Object Detection}

\author{Qianqian Zhang,
    Leon Tabaro,
	Ahmed M. Abdelmoniem,
	and Junshe An

\thanks{This work was supported in part by the China Scholarship Council (CSC) program (Project ID: 202504910309) and Research on the Working Principle and Application of Intelligent Satellite Brain under the National Key Research and Development Program of China (Project ID: 2022YFF0503900) and UKRI EPSRC Grant Reference EP/X035085/1. This work was completed entirely while Qianqian Zhang was a visiting researcher at Queen Mary University of London, UK. (Corresponding~authors: Qianqian Zhang)

Qianqian Zhang is with the National Space Science Center, Chinese Academy of Sciences, Beijing, 101499, China, the School of Computer Science and Technology, University of Chinese Academy of Sciences, Beijing, 100049, China, and the School of Electronic Engineering and Computer Science, Queen Mary University of London, London, E1 4NS, UK (e-mail: zhangqianqian21@mails.ucas.ac.cn).

Leon Tabaro and Ahmed M. Abdelmoniem are with the School of Electronic Engineering and Computer Science, Queen Mary University of London, London, E1 4NS, UK (e-mails: {l.tabaro@qmul.ac.uk, ahmed.sayed@qmul.ac.uk}).

%Ahmed M. Abdelmoniem is with the School of Electronic Engineering and Computer Science, Queen Mary University of London, London, E1 4NS, UK (e-mail: ahmed.sayed@qmul.ac.uk).

Junshe An is with the National Space Science Center, Chinese Academy of Sciences, Beijing, 101499, China, and the School of Astronomy and Space Science, University of Chinese Academy of Sciences, Beijing, 100049, China (e-mail: anjunshe@nssc.ac.cn).

}
}
%倩倩todo 在下面添加标题替换
\markboth{IEEE TRANSACTIONS ON GEOSCIENCE AND REMOTE SENSING,~Vol.~X, No.~X, 2026}%
{Zhang \MakeLowercase{\textit{et al.}}: XXXXXXX}

\maketitle

% !TeX spellcheck = en_US
%倩倩todo 完成摘要
\begin{abstract}  
%%%%%%%%%qianqian todo
% Task
% Technical challenge for previous methods (围绕我们解决了的technical challenge展开讨论)
% 一两句话介绍解决challenge的technical contribution (一般就提到xxx技术的名词，不会讲具体的每个步骤。这个名词要让人读得懂，不要有jump的感觉。这个能力对写好abstract很重要。)
% 介绍technical contribution的好处
% Experiment
Multispectral fusion object detection is a critical task for edge-based maritime surveillance and remote sensing, demanding both high inference efficiency and robust feature representation for high-resolution inputs. However, current State Space Models (SSMs) like Mamba suffer from significant parameter redundancy in their standard 2D Selective Scan (SS2D) blocks, which hinders deployment on resource-constrained hardware and leads to the loss of fine-grained structural information during conventional compression. To address these challenges, we propose the Low-Rank Two-Dimensional Selective Structured State Space Model (Low-Rank SS2D), which reformulates state transitions via matrix factorization to exploit intrinsic feature sparsity. Furthermore, we introduce a Structure-Aware Distillation strategy that aligns the internal latent state dynamics of the student with a full-rank teacher model to compensate for potential representation degradation. This approach substantially reduces computational complexity and memory footprint while preserving the high-fidelity spatial modeling required for object recognition. Extensive experiments on five benchmark datasets and real-world edge platforms, such as Raspberry Pi 5, demonstrate that our method achieves a superior efficiency-accuracy trade-off, significantly outperforming existing lightweight architectures in practical deployment scenarios.

% Multispectral fusion small target detection is a critical task in real-world applications including maritime surveillance, remote sensing, and urban security. The practical deployment of such tasks typically requires execution on resource-constrained edge devices. However, existing State Space Model based detectors, particularly those built upon Mamba with standard SS2D blocks, suffer from severe parameter redundancy and inefficient spatial dependency modeling, which limits their deployment on high-resolution inputs and edge devices. To address this challenge, we propose a Low-Rank Two-Dimensional Selective Structured State Space Model that reformulates state transitions via matrix factorization to exploit the intrinsic low-rank property of visual features. Furthermore, we introduce a Structure-Aware Distillation strategy that aligns singular components and hidden state dynamics between full-rank and low-rank models to compensate for representational degradation. The proposed framework significantly reduces parameters and computational complexity while preserving long-range spatial modeling capability and discriminative power for small targets. Extensive experiments on five benchmarks and cross-platform evaluations on NVIDIA A100, RTX 4090, and Raspberry Pi 5 demonstrate that our method achieves competitive accuracy with substantially improved real-time inference efficiency on both high-end and edge hardware.
%%%%%%%%%qianqian todo

\end{abstract}

%倩倩todo  根据边缘资源受限完成以下关键词
% !TeX spellcheck = en_US
\section{Introduction}

%%%%%%%%%qianqian todo
% Task and application
% Technical challenge for previous methods (围绕我们解决了的technical challenge展开讨论。Technical challenge包括limitation和technical reason)
% 介绍解决challenge的our pipeline
% Experiment
% Contributions
%%%%%%%%%%%qianqian todo

%%%%%%%%%%%qianqian todo  写完intro第一自然段%%%%
%段落写作：一段文字开头第一句就要让人知道这段在说什么，并且一段文字一定要表达好一件事情。【写完自己检查，是否一段只说一件事，段首句为全部】
% 【todo1：】Task and application
%【todo2：】通过介绍previous methods来引出想解决的technical challenge（想解决的failure cases、想提升的任务指标）
\IEEEPARstart{O}{bject} detection plays an important role in various fields, including maritime surveillance \cite{chen2024weather}, remote sensing \cite{wen2023comprehensive}, and urban security tasks \cite{izquierdo2025strategic}. In these scenarios, objects are highly susceptible to environmental noise and variations in illumination conditions. While traditional single-spectral detection methods have established a solid foundation, they frequently suffer performance degradation in complex environments \cite{10075555}. To address this, fusing multi-spectral information has emerged as a robust alternative, as it effectively complements object features by integrating diverse physical properties, such as thermal signatures and textural details \cite{zhang2019cross, zhang2021guided, 10570450}.

The practical deployment of such tasks typically requires execution on resource-constrained edge devices. Moreover, with the rapid advancement of imaging sensors, the resolution of input images has increased significantly, demanding higher inference efficiency \cite{10075555, wu2025contour}. Compared to Convolutional Neural Networks (CNNs) and Vision Transformers (ViTs), the recently proposed State Space Model (Mamba) architecture \cite{gu2023mamba, liu2024vmamba}, has demonstrated a superior ability to maintain long-range spatial modeling while achieving linear computational complexity. This characteristic makes Mamba an ideal backbone for processing high-resolution images on efficiency-sensitive platforms.

However, we observe that standard 2D Selective Scan (SS2D) blocks exhibit significant technical limitations when handling dense spatial dependencies. Specifically, the standard SS2D layers often exhibit substantial parameter redundancy, which severely limits their deployment on resource-constrained edge hardware. A well-established solution to this redundancy is currently missing because existing compression techniques often fail to preserve the fine-grained structural information essential for object detection. The technical reason lies in the misalignment between parameter reduction and the need for high-fidelity spatial representation in SSMs.

In this paper, we are motivated to develop a more efficient recognition framework that compresses structural information without sacrificing discriminative power. 

%【done，需要润色】%%%%%%%%%%qianqian todo  写完intro第3自然段%%%%
% 介绍解决challenge的our pipeline，method1

\begin{figure}[htbp]
	\centering
	%\color{red}
	\includegraphics[width=0.5\textwidth]{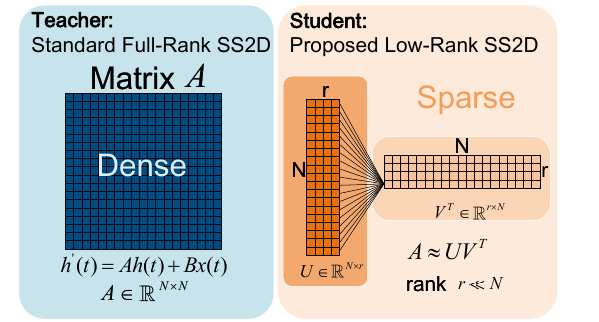}
	\centering
	\caption{Structural comparison of full-rank vs. low-rank SS2D. By significantly reducing computational overhead while maintaining representative power, the low-rank design opens up new avenues for efficient vision computing on resource-constrained edge devices.}
	\label{teaser}
    \vspace{-6pt}
\end{figure}

% %【Todo：需要润色，逐点check以下四点】
% (1) 我们的pipeline解决了什么technical challenge。
% (2) 我们的technical contribution是什么。
% (3) 我们方法本质上能work的原因是什么。
% (4) 我们方法相对于之前方法的好处是什么。
\noindent{\textbf{Low Rank Two Dimensional Selective Structured State Space Model for Efficient Recognition.}} Traditional state space models such as Mamba encounter significant technical hurdles in visual tasks, specifically excessive computational redundancy and overparameterized structures that impede deployment on resource-constrained edge devices. To address these limitations, we propose a Low-Rank two dimensional state space (SS2D) modeling approach, termed Low Rank SS2D, specifically tailored for visual recognition. Instead of constructing state transitions using full rank matrices, the core transition component is reformulated via matrix factorization. As shown in Fig.~\ref{teaser}. The primary technical advantage of this method lies in its ability to exploit the intrinsic sparsity and low-rank characteristics of visual features. Consequently, the proposed architecture substantially reduces model parameters and computational complexity while preserving the capacity to model long-range spatial dependencies. 

%【done，需要润色】%%%%%%%%%%qianqian todo  写完intro第3自然段%%%%

%【done，需要润色】%%%%%%%%%%qianqian todo  写完intro第4自然段%%%%
% 介绍解决challenge的our pipeline，method2
% %【Todo：需要润色，逐点check以下四点】
% (1) 我们的pipeline解决了什么technical challenge。
% (2) 我们的technical contribution是什么。
% (3) 我们方法本质上能work的原因是什么。
% (4) 我们方法相对于之前方法的好处是什么。
\noindent{\textbf{Structure Aware Distillation Tailored for Low Rank SS2D.}}
To mitigate the potential degradation in representational capacity induced by low-rank compression, we introduce a structure-aware distillation strategy for the Low Rank SS2D framework. Unlike conventional distillation methods, our approach leverages a full-rank teacher model to guide the low-rank student model via a multidimensional loss function. This objective comprises singular value decomposition (SVD)- aligned distillation, hidden-state sequence distillation, and feature-reconstruction distillation. The technical advantage of this method is its ability to precisely compensate for information loss caused by low-rank modeling. By compelling the student model to emulate the teacher's internal state-transition dynamics rather than merely matching final outputs, this paradigm extends beyond simple knowledge compression. Aligning the internal latent state spaces enables lightweight models to reproduce the complex spatiotemporal reasoning capabilities of large-scale models.
%【done，需要润色】%%%%%%%%%%qianqian todo  写完intro第4自然段%%%%

%【done，需要润色】%%%%%%%%%%qianqian todo  写完intro第5自然段%%%%
% Experiment
\noindent{\textbf{Comprehensive Cross Platform Validation with Real World Edge Deployment.}} Existing lightweight visual models often emphasize theoretical reductions in computational complexity but lack empirical validation of real-world inference latency across heterogeneous platforms. We conduct extensive experiments on five benchmark datasets to validate the effectiveness of the proposed approach. In addition to measuring real-time inference performance on high-end GPUs, including the NVIDIA A100 and NVIDIA GeForce RTX 4090, we implement and benchmark deployment on edge platforms such as the Raspberry Pi 5. Experimental results demonstrate that our method maintains high recognition accuracy while achieving efficient inference on computationally constrained edge devices. These findings validate the practical applicability of the proposed framework in real-world scenarios.

%【done，需要润色】%%%%%%%%%%qianqian todo  写完intro第5自然段%%%%

%%%%%%%%%%%qianqian todo  写完intro部分的contribution%%%%
In summary, this article makes the following contributions.
\begin{itemize}
	\item We propose a novel Low Rank SS2D architecture that significantly reduces computational redundancy while preserving the ability to model long-range spatial dependencies, enabling efficient visual recognition on edge devices.
	\item We introduce a structure aware distillation strategy tailored for low rank models, which effectively compensates for information loss and allows lightweight models to replicate the complex reasoning capabilities of larger models.
	\item We conduct comprehensive experiments across multiple benchmark datasets and real world edge platforms, demonstrating that our method achieves high accuracy while maintaining efficient inference, validating its practical applicability in real world scenarios.
	\item To the best of our knowledge, this is among the few works to systematically address the challenges of deploying state space models for visual recognition on resource-constrained edge devices, providing a new paradigm for efficient model design and deployment in this domain.
\end{itemize}

%%%%%%%%%%%qianqian todo  写完intro部分的contribution%%%%
% !TeX spellcheck = en_US
\section{Related Work}
\label{sec:Related Work}
%%%%%%%%%qianqian todo
% 要写出好的Related work，步骤为：
% (1) 首先，列出和自己论文的方法比较相关的论文。【todo：需要找2024年-2026年最新的，以及应当提及result章节实验表格中的方法】（Related work中最重要的部分，如果没讨论，一些reviewer会直接以此拒掉论文）
% (2) 然后，根据论文的研究方向和算法技术来确定Related work要讨论哪几个topics【已完成】，列出几个topics下面要讨论的论文【todo：需要根据找到的2024年-2026年的最新内容，以及应当提及result章节实验表格中的方法，分类，写】
% (3) 最后，基于前两步列出的论文组织related work的写作思路。【todo：待列出写作思路】

% 然后要写出它们的方法的局限性在哪里，而你的认为可能可以解决的方法是XX，最后引出你的方法。【todo：待写】

%% For Leon:段落写作：一段文字开头第一句就要让人知道这段在说什么，并且一段文字一定要表达好一件事情。【For Leon:每段第一句要说清楚这段说的主要内容。】
% For Leon: Paragraph Writing: The first sentence of each paragraph should clearly inform the reader what the paragraph is about, and each paragraph must effectively convey one specific idea. 【For Leon: The first sentence of each paragraph should clearly state the main content of the paragraph.】
%%%%%%%%%qianqian todo

% The emergence of deep learning has marked a profound shift in the paradigm of computer vision and has in recent decades driven remarkable progress in object detection for natural scene images, such as face recognition and pedestrian detection \cite{wu2025contour, CAO2026131878}. However, unlike natural scenes, 
% By leveraging complementary information from different spectral bands, particularly visible (RGB) and infrared (IR), multispectral detection frameworks aim to overcome the unique challenges posed by single-modal data in RSI applications.\\ 

\subsection{Multispectral Fusion for Object Detection}
%重点： 综述多光谱（红外+可见光）融合的现状。讨论传统方法与深度学习方法在处理小目标（特征缺失、背景复杂）时的局限性，引出融合的必要性。
%% For Leon:段落写作：一段文字开头第一句就要让人知道这段在说什么，并且一段文字一定要表达好一件事情。
% For Leon: Paragraph Writing: The first sentence of each paragraph should clearly inform the reader what the paragraph is about, and each paragraph must effectively convey one specific idea. 【For Leon: The first sentence of each paragraph should clearly state the main content of the paragraph.】

Recently, in the field of remote sensing imaging (RSI), multispectral fusion has attracted increasing attention for improving object detection performance under complex environmental conditions. Unlike natural scene images, remote sensing data often exhibit large-scale variations, occlusion, and severe background interference \cite{hua2025survey}. While recent single-modal RGB detectors have advanced through multiscale feature reconstruction and dynamic receptive field modeling \cite{11300300, zhang2025lga, wang2025position} their performance remains inherently fragile and suffers from significant degradation under variations in illumination, nighttime, or adverse weather conditions. In contrast, infrared (IR) images captures thermal radiation patterns that are largely invariant to lighting conditions and can highlight object contours even in challenging environments, yet generally exhibits lower spatial resolution, reduced texture detail, and limited semantic richness compared to RGB images \cite{10757439}. These complementary characteristics naturally motivate the integration of visible and infrared modalities for more reliable object detection in RSI technologies \cite{tang2023rethinking, shen2024icafusion,yuan2024c2former, hu2025ei}.\\
To realize effective fusion, early deep learning-based approaches predominantly employ convolutional neural networks (CNNs) to extract hierarchical spatial features and fine-grained local patterns. SuperYOLO \cite{10075555} introduces a pixel-level symmetric multimodal fusion module to efficiently combine RGB and IR inputs. However, the locality bias induced by the fixed receptive field size in CNN-based methods limits their ability to learn global contextual information. And so to overcome these limitations, several hybrid architectures combining convolutional backbones with attention-based interaction modules have been proposed to explicitly capture long-range dependencies and address the challenges of intermodal discrepancies and intramodal variability in RGB-IR fusion \cite{zhu2024cross, 11180153, li2025crossmodalnet, qi2026small, yue2025diffusion}.\\ 
Unfortunately, in spite of the ubiquity and strong performance of Transformer architectures in modelling long-range dependencies, the $\mathcal{O}(N^2)$ complexity of the attention mechanism is extremely wasteful, and becomes a significant bottleneck for many applications involving long sequences such as high-resolution sensing images \cite{gu2023mamba}. This redundancy results in both high memory consumption and high inference latency, whereas systems for such real-world applications often demand real-time responses with low latency and inference cost.

% Although multiscale pyramids partially alleviate this issue \cite{sapkota2025yolo}, convolutional operators remain fundamentally constrained in modeling long-range dependencies, critical for handling high-resolution remote sensing images.

% CrossModalNET \cite{li2025crossmodalnet} introduces a bidirectional cross-modal attention at the channel level to adaptively align visible/IR features while mitigating information loss under strong cross modality discrepancies. MASSA \cite{qi2026small}, further replaces SuperYOLO’s simple pixel-level symmetric fusion which fails to model channel wise importance causing redundancy by diluting critical IR thermal vs RGB signals, with a multi-scale self-attention aggregation mechanism, enabling adaptive channel-wise weighting across scales during fusion.

\subsection{Vision Mamba for Efficient Visual Representation}
%重点： 聚焦 Mamba (SSM) 及其在视觉任务中的应用。对比 Transformer 的计算复杂度，强调 Mamba 在保持长距离建模能力的同时，如何提高推理效率，从而引出你的架构基础。
%% For Leon:段落写作：一段文字开头第一句就要让人知道这段在说什么，并且一段文字一定要表达好一件事情。
% For Leon: Paragraph Writing: The first sentence of each paragraph should clearly inform the reader what the paragraph is about, and each paragraph must effectively convey one specific idea. 【For Leon: The first sentence of each paragraph should clearly state the main content of the paragraph.】
Recently, Mamba \cite{gu2023mamba}, built on Structured State-Space Sequence Models (S4) \cite{gu2022efficiently}, has emerged as a promising class of architectures for sequence modelling that addresses the inefficiency of CNNs and Transformers. By maintaining a fixed-size hidden-state representation, the Mamba model scales linearly with the input sequence length while preserving comparable expressive capacity through an adaptive selective-scan mechanism. VMamba \cite{liu2024vmamba} extends this framework to visual tasks through the 2D Selective Scan (SS2D) mechanism, effectively capturing long-range spatial dependencies.
Recently, Vision Mamba \cite{zhu2024vision}, and SS2D-based architectures have been progressively introduced into multispectral remote sensing tasks \cite{li2025multispectral,zhou2025dmm, feng2025mvmamba}. DMM \cite{zhou2025dmm} introduces a disparity-guided selective scanning to mitigate intermodal discrepancies in oriented remote sensing detection. More recently, MVMamba \cite{feng2025mvmamba} integrates state-space duality (SSD) mechanisms with multiscale fusion to enhance small-object detection performance in high-resolution remote sensing imagery.\\
Despite these advances, existing Mamba-based detection frameworks primarily focus on improving representation capability and cross-modal fusion performance, whereas comparatively less attention has been paid to the structural efficiency of the Mamba operator when deployed on resource-constrained edge devices such as smart satellites and drones, which often have limited memory and processing power. This work departs from prior efforts by shifting the focus to developing a compact yet expressive state-space modeling framework tailored for efficient multispectral object detection under edge-deployment conditions.

% For example, MRL-Mamba \cite{li2025multi} and UAV-Mamba \cite{yu2025uav} exploit Mamba’s linear-complexity global modeling to enhance long-range contextual reasoning for infrared small target detection under cluttered backgrounds and real-time UAV constraints. Similarly, MGMamba \cite{xiao2025multi} extends the selective scanning mechanism to multi-directional grouped representations to better capture spatial dependencies in dense infrared scenes.
\vspace{-4pt}
\subsection{Model Compression and Knowledge Distillation for Edge Deployment}
%重点： 探讨如何将大型模型部署到边缘端。讨论低秩分解（Low-Rank）和知识蒸馏（Distilling）的技术演进，为ours轻量化策略提供背景支持。
%% For Leon:段落写作：一段文字开头第一句就要让人知道这段在说什么，并且一段文字一定要表达好一件事情。
% For Leon: Paragraph Writing: The first sentence of each paragraph should clearly inform the reader what the paragraph is about, and each paragraph must effectively convey one specific idea. 【For Leon: The first sentence of each paragraph should clearly state the main content of the paragraph.】

While modern deep neural networks (DNNs) have significantly advanced aerial image analysis, their high computational demands can make them prohibitive for deployment on resource-limited devices such as mobile phones and embedded systems. To mitigate this challenge, a long-standing strategy has been to exploit the redundancy of overparameterized networks by replacing dense linear transformations with low-rank factorizations \cite{huh2103low, kwon2024efficient}. 
\subsubsection{Low Rank Decomposition}
Formally, low-rank decomposition factorizes a large matrix $W \in \mathbb{R}^{m \times n}$ into two smaller matrices $U \in \mathbb{R}^{m \times k}$ and  $V \in \mathbb{R}^{k \times n}$, such that  $W = UV$, where $k \ll m$ and $k \ll n$. This reduces the parameter complexity from $mn$ to $k(m+n)$. Such techniques have been extensively studied for compressing various components of neural architectures and, in recent years, have been widely adopted for compressing large pretrained models \cite{li2023losparse, hsu2022language}, thereby demonstrating significant reductions in computational and memory footprints. 

Moreover, recent theoretical and empirical studies have revealed that deep networks exhibit intrinsic spectral concentration and rank diminishing behavior across layers, leading to low-dimensional feature representations \cite{huh2103low, feng2022rank}. This suggests that much of the expressive capacity of overparameterized models is concentrated in a small set of dominant singular directions.

\subsubsection{Knowledge Distillation}
Knowledge Distillation (KD) has emerged as a pivotal technique for deploying large-scale models onto edge devices without substantial performance degradation \cite{hinton2015distilling}. Established KD methodologies are generally categorized into response-based, feature-based, and relation-based distillation \cite{moslemi2024survey}. Notably, knowledge distillation has been widely applied in remote sensing tasks to improve the performance of lightweight detection models \cite{yue2025diffusion, zeng2024novel, xue2024feature, huang2024optimizing, zhang2025knowledge}. IRKD \cite{xue2024feature} proposes a feature-based distillation framework tailored for infrared small-target detection, in which channel–spatial attention masks selectively transfer informative features from the teacher to the student network. Closely related to this work, TDKD-Net \cite{zeng2024novel} introduces a tensor decomposition-based compression strategy for UAV object detection and employs response-based distillation to compensate for accuracy degradation caused by low-rank convolutional approximation.\\ 

\section{Baseline Architecture}
\label{sec:Baseline}

%%%%%%%%%qianqian todo
% 
The baseline framework mainly consists of three functional components: the Pixel-level Multi-modal Fusion Module, the State Space 2D Backbone, and the Detection Network with a YOLOv8n\cite{yolov8_ultralytics} Head. In the subsequent subsections, we first give a brief introduction to the SS2D mechanism, then elaborate on each component of the baseline in detail.

\subsection{Preliminaries: SS2D Mechanism} 
\label{subsec:SS2D}

As outlined in the baseline framework, the State Space 2D (SS2D) backbone serves as the core for feature extraction. This subsection first reviews the foundational State Space Models (SSM), specifically Mamba~\cite{gu2023mamba}, and then elaborates on how VMamba~\cite{liu2024vmamba} extends this mechanism to 2D visual data through the SS2D module.

\subsubsection{Mamba: Selective State Space Modeling}
SSMs map an input sequence $x(t) \in \mathbb{R}$ to an output $y(t) \in \mathbb{R}$ via a hidden state $h(t) \in \mathbb{R}^{\mathtt{N}}$. In a continuous-time 1-D context, the system is defined by linear ordinary differential equations (ODEs):

\begin{equation}
\begin{aligned}
\label{eq:lti}
h'(t) &= \mathbf{A}h(t) + \mathbf{B}x(t), \\
y(t) &= \mathbf{C}h(t),
\end{aligned}
\end{equation}

where $\mathbf{A} \in \mathbb{R}^{\mathtt{N} \times \mathtt{N}}$ is the state transition matrix, $\mathbf{B} \in \mathbb{R}^{\mathtt{N} \times 1}$ is the input coefficient vector, and $\mathbf{C} \in \mathbb{R}^{1 \times \mathtt{N}}$ serves as the output coefficient vector.

To handle discrete sequences, Mamba~\cite{gu2023mamba} introduces a selective mechanism by making these matrices dynamically dependent on the input. It discretizes the system using a time step parameter $\mathbf{\Delta}$ via the zero-order hold (ZOH) method:

\begin{equation}
\begin{aligned}
\label{eq:zoh}
\mathbf{\overline{A}} &= \exp{(\mathbf{\Delta}\mathbf{A})}, \\
\mathbf{\overline{B}} &= (\mathbf{\Delta} \mathbf{A})^{-1}(\exp{(\mathbf{\Delta} \mathbf{A})} - \mathbf{I}) \cdot \mathbf{\Delta} \mathbf{B}.
\end{aligned}
\end{equation}

After discretization, the recurrence is formulated as:

\begin{equation}
\begin{aligned}
\label{eq:discrete_lti}
h_k &= \mathbf{\overline{A}}h_{k-1} + \mathbf{\overline{B}}x_{k}, \\
y_k &= \mathbf{C}h_k.
\end{aligned}
\end{equation}

This design ensures linear complexity and efficient training; however, its sequential scanning is inherently 1-D, making it suboptimal for 2-D spatial data like images, which lack a predefined scan order.

\subsubsection{VMamba and SS2D Module}
\begin{figure}[htbp]
	\centering
	%\color{red}
	\includegraphics[width=0.48\textwidth]{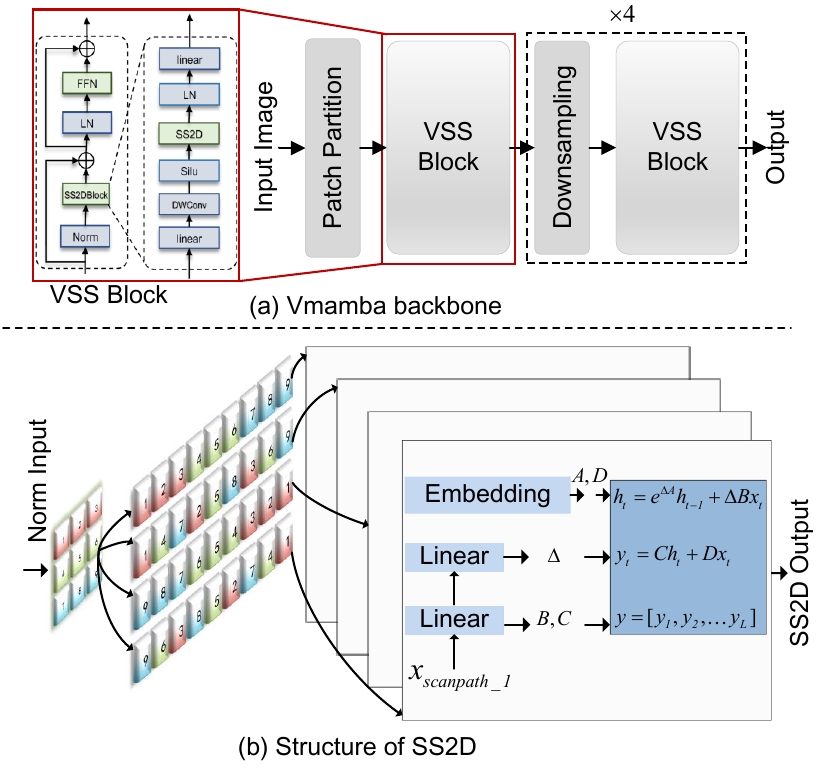}
	\centering
	\caption{Overview of the VMamba backbone and SS2D module. }
	\label{ss2d}
    \vspace{-6pt}
\end{figure}

To bridge the gap between 1-D scanning and 2-D spatial information, VMamba~\cite{liu2024vmamba} introduces the Visual State Space (VSS) block. Its core component, the SS2D module, addresses the sequential scanning problem by traversing images across four different scan paths as shown in Fig.~\ref{ss2d} (b). This ensures that each pixel captures dependencies from multiple directions, effectively modeling local textures and global structures.

Furthermore, the VSS block enhances computational efficiency by utilizing a structure of network branches and twin residual modules instead of traditional multiplicative branches, Fig.~\ref{ss2d} (a). 

\vspace{-4pt}
\subsection{Pixel-level Multi-modal Fusion} 
\label{subsec:fusion}
 A primary challenge in multi-modal object detection is the environmental sensitivity of individual sensors: RGB images exhibit degraded performance in low-light conditions, whereas infrared (IR) images lack rich texture information. Existing methods typically perform fusion at the deep feature level, which may lead to the loss of fine-grained spatial information. To address this issue, we draw on the pixel-level fusion module proposed in \cite{zhang2025selective} to construct a unified and robust input representation at the earliest stage of the network (As shown in Fig.~\ref{overview}). The Pixel-level Multi-modal Fusion module integrates complementary information from different sensors to generate a unified representation $I^{f}$. The detailed workflow is summarized as follows:

Given a pair of visible--infrared images, we denote the visible image as 
$I^{v} \in \mathbb{R}^{H \times W \times C}$ and the infrared image as 
$I^{i} \in \mathbb{R}^{H \times W \times C}$, where $H$, $W$, and $C$ represent 
the height, width, and number of channels of the input images, respectively.

The two modalities are fused at the pixel level through the operator 
$\mathcal{F}_{\text{fusion}}$.
%  Given a pair of spatially aligned images, we denote the visible light image as $I^{v} \in \mathbb{R}^{H \times W \times C}$ and the infrared image as $I^{i} \in \mathbb{R}^{H \times W \times C}$, where $H$, $W$, and $C$ represent the height, width, and channel dimension of the input images, respectively.
 
% We apply a pixel-wise fusion operator $\mathcal{F}_{\text{fusion}}$, which first conducts local weighted aggregation or concatenation of the two modalities, followed by a convolution layer to align the channel dimensions.

The fused multi-modal representation $I^{f} \in \mathbb{R}^{H \times W \times C_{\text{in}}}$ is produced, where $C_{\text{in}}$ denotes the channel depth of the fused feature map.

Mathematically, the fusion operation is formulated as:
\begin{equation}
    \label{eq:fusion}
    I^{f} = \mathcal{F}_{\text{fusion}}(I^{v}, I^{i})
\end{equation}
By performing pixel-level fusion, fine-grained details critical for object detection can be preserved, thereby significantly enhancing the model’s robustness under extreme illumination variations and sensor noise.

\section{The Proposed Framework}
\label{sec:Ours}
\begin{figure*}[htbp]
	\centering
	%\color{red}
	\includegraphics[width=1\textwidth]{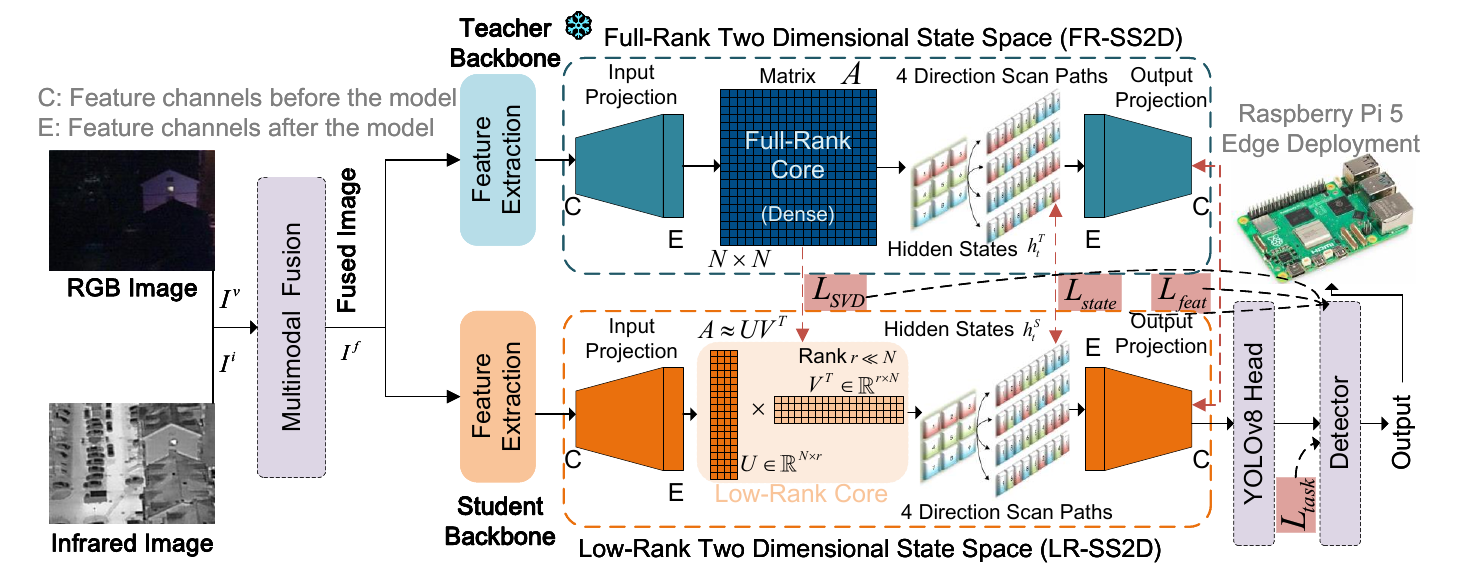}
	\centering
	\caption{Overview of the proposed DLRMamba framework. The proposed framework consists of four core components: (1) A pixel-level multispectral modality fusion module, which is designed to effectively fuse and process visible and infrared spectral information; (2) Low-Rank Structured State Space Modeling (Low-Rank SS2D), which is integrated to realize model lightweighting; (3) A structure-aware distillation (SAD) mechanism, including Singular Value Decomposition (SVD) Alignment (Matrix-level Distillation), Hidden State Sequence Alignment (Dynamic Distillation), and Feature Reconstruction (Output-level Distillation), which is proposed to compensate for performance degradation induced by model compression; (4) A detection head, which is used to output the final detection results.}
	\label{overview}
    \vspace{-6pt}
\end{figure*}

The proposed framework aims to achieve efficient multi-modal object detection by compressing the two-dimensional state space model (SS2D) while maintaining high-fidelity feature representation. While leveraging the baseline's pixel-level multi-modal fusion and task-specific Detection Head, the framework introduces two pivotal enhancements: Low-Rank SS2D Representation and Structure-Aware Distillation, ensuring high-fidelity feature extraction despite the reduced model footprint. As shown in Fig.~\ref{overview}.

\vspace{-4pt}
\subsection{Low-Rank 2D Selective Structured State Space Model} 
\label{subsec:1}

%(2) 对于每一个模块，回答三个问题：这个module的工作流程、为什么要用这个
A remaining bottleneck in deploying SS2D on edge devices is the high computational complexity of the full-rank system matrix $A$. In standard SS2D, the hidden state transition entails dense matrix operations that pose significant challenges for real-time inference. To mitigate this, we introduce the Low-Rank SS2D module, which leverages low-rank factorization to reduce the parameter count while preserving a large global receptive field.

The Low-Rank SS2D module replaces the standard full-rank state transition with a factorized alternative. The detailed design is as follows:

Based on the Singular Value Decomposition (SVD) property, the full-rank system matrix $A \in \mathbb{R}^{N \times N}$ is decomposed into two low-rank matrices $U \in \mathbb{R}^{N \times r}$ and $V \in \mathbb{R}^{N \times r}$, where $r \ll N$.

Instead of computing $A h_{t-1}$, the student model computes the transition in two smaller steps: first projecting the previous state $h_{t-1}$ into a low-dimensional subspace via $V^T$, and then projecting it back via $U$.

The discretized state-space equation for the student model is updated as:
    \begin{equation}
    h_t^{s} = (U V^T) h_{t-1}^{s} + B x_t
    \end{equation}

This low-rank structure significantly accelerates inference speed on resource-constrained hardware while maintaining the capability to model long-range dependencies inherent in the SS2D architecture.
\vspace{-4pt}
\subsection{Structure-Aware Distillation Tailored for Low-Rank SS2D} 
\label{subsec:17}

%(2) 对于每一个模块，回答三个问题：这个module的工作流程、为什么要用这个
Simply reducing the rank of the SS2D module results in a performance gap relative to the full-rank teacher model. The key challenge lies in compensating for information loss during compression. We propose a Structure-Aware Distillation module to guide the student model in inheriting the teacher’s internal spatio-temporal dynamics and weight structures. 

The Structure-Aware Distillation module utilizes a triple-alignment strategy to supervise the student model. The detailed implementation steps are as follows:

\subsubsection{SVD Alignment (Matrix-level Distillation)} 

We align the student's low-rank matrices $U_{s}, V_{s}$ with the principal singular components of the teacher's matrix $A_{t}$. Specifically, we denote the aligned principal singular components of the teacher as $U_{t}$ and $V_{t}$, The corresponding loss is then defined as:
\begin{equation}
\mathcal{L}_{SVD} = \left\| U_{s} - U_{t} \right\|_F^2 + \left\| V_{s} - V_{t} \right\|_F^2
\end{equation}

where $U_{t} = (U_t \sqrt{\Sigma_t})_{1:r}$ and $V_{t} = (V_t \sqrt{\Sigma_t})_{1:r}$.

\subsubsection{Hidden state sequence Alignment (Dynamic Distillation)} This is the key to capturing long-range dependencies. SS2D generates a series of hidden states 
$H = \{h_1, h_2, \dots, h_L\}$ when scanning an image. 
We require the state trajectory $H_{\text{stud}}$ of the Student to mimic the state trajectory 
$H_{\text{teach}}$ of the Teacher, ensuring that both respond consistently to dynamic features:
\begin{equation}
\mathcal{L}_{\text{state}} = \frac{1}{L} \sum_{t=1}^{L} \text{MSE}\left(h_t^{\text{s}}, P\left(h_t^{\text{t}}\right)\right)
\end{equation}

where $P$ is a dimension-adaptive projection layer used to align the potentially different state dimensions of the Teacher and the Student. MSE is the Mean Squared Error, which measures the average squared difference between the student and teacher hidden states across all time steps. By minimizing this loss, we encourage the student model to closely follow the temporal dynamics of the teacher model, thus preserving the long-range dependencies that are crucial for accurate feature representation in SS2D.

\subsubsection{Feature reconstruction
 (Output-level Distillation)} To ensure the semantic consistency of the final feature map, we perform reconstruction distillation on the output $Y$ of the SS2D module. We minimize the distance between the output feature maps $Y_{t}$ and $Y_{s}$ to guarantee semantic consistency, with the loss defined as: 
\begin{equation}
 \mathcal{L}_{feat} = \left\| Y_{t} - Y_{s} \right\|_2^2
\end{equation}

 Structure-Aware Distillation, specifically tailored to the Low-Rank SS2D module, ensures structural fidelity during compression. By aligning hidden-state trajectories and weight-decomposition manifolds, the student model effectively captures the teacher's fine-grained reasoning logic. This multi-dimensional supervision mechanism bridges the performance gap typically introduced by low-rank approximations.

\subsection{Task Specific Detector}
After extracting efficient multi-modal features, the final challenge lies in translating these representations into accurate object localizations and classifications. To address this, we employ a Decoupled Detection Head to process the task.

The Detection Head processes the multi-scale feature pyramid generated by the Low-Rank SS2D backbone. The workflow is as follows:

First, the module takes as input multi-scale features denoted as $P = \{P_3, P_4, P_5\}$, where each element corresponds to a feature map at a different spatial scale; second, each feature map is fed into two parallel branches, with one branch dedicated to Bounding Box Regression and the other to Class Probability prediction; finally, the head outputs the final detection results $\hat{Y}_{det} = \{ \hat{B}_{box}, \hat{C}_{cls} \}$, encompassing both bounding box coordinates and class probabilities.

The total training objective is formulated as the combination of the primary detection task loss and auxiliary distillation losses:
\begin{equation}
\mathcal{L}_{Total} = \lambda_1 \mathcal{L}_{Task}(\hat{Y}_{det}, GT) + \lambda_2 \mathcal{L}_{SVD} + \lambda_3 \mathcal{L}_{state} + \lambda_4 \mathcal{L}_{feat}
\label{loss}
\end{equation}

where $\lambda_1=1.0$, $\lambda_2=0.5$, $\lambda_3=0.1$, and $\lambda_4=1.5$ denote the weighting coefficients for the corresponding losses, respectively.

The decoupled detection design facilitates faster convergence and enhances both precision and recall. Furthermore, integrating distillation losses into the overall objective function enables the detection head to leverage highly distilled, efficient features, ultimately yielding superior performance on devices.
\vspace{-8pt}

\section{Experimental Results}
\label{sec:Experiment}
In this section, we introduce the five datasets used for evaluation, the implementation details, the accuracy metrics, as well as extensive comparison experiments and ablation studies to verify the effectiveness of the proposed method.
%%%%%%%%%qianqian todo
% 要写出好的Experiments，需要回答三个问题：
% (1) 怎么证明我们的方法比已有方法更强 → 做哪些comparison experiments。
% (2) 怎么证明方法里的module有效 → 做哪些ablation studies。
% (3) 怎么充分展示我们方法的上限 → 在哪些更有挑战性的数据上做demo。

% Experiments的文字部分，比较重要的是figure和table的caption。

% Table caption、Figure caption里需要写清楚experimental setting、notation。如果没啥好说的，可以一句话简单说一下实验结果。
% Caption的内容不要大篇幅地讨论实验结果，这样容易和正文重复。

% 实验图表的排版技巧：单栏的图/表，放在论文的右栏比较好看，因为人的阅读习惯会从左上角找第一行文字。

% - 做哪些ablation studies
    
%     一篇论文包含了一些core contributions和每个pipeline module中的一些design choices。读者通常会**很在意core contributions对performance的影响**，并且会**好奇这些design choices是否真的有用**。
    
%     因此，ablation studies通常需要包含两个部分：
    
%     1. 一个大表和相应的可视化对比图，列出论文的core contributions以及一些重要的components对论文方法performance的影响。
%     2.一些小表和相应的可视化对比图。每个小表分别列出一个pipeline module中design choices对论文方法performance的影响（方法对超参的敏感性，方法对input data质量的敏感性，不采用某个design choice对performance的影响）。

%%%%%%%%%qianqian todo

\subsection{Dataset}

\begin{figure}[t]
	\centering
	%\color{red}
	\includegraphics[width=0.48\textwidth]{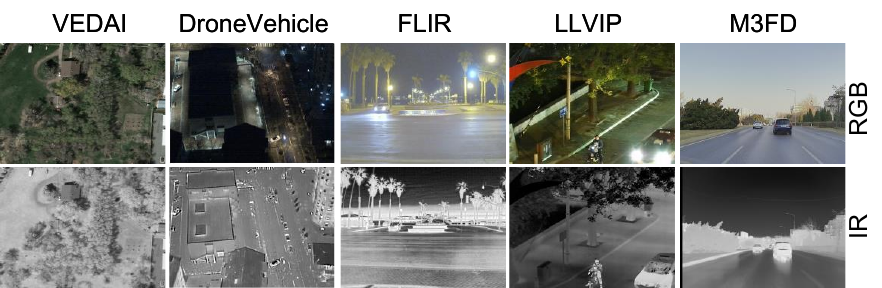}
	\centering
	\caption{Sample RGB–IR pairs from five datasets (top: RGB; bottom: IR).}
	\label{example}
    \vspace{-6pt}
\end{figure}

We evaluate our method on five widely used RGB-IR object detection datasets: VEDAI~\cite{Razakarivony2016Vehicle}, FLIR~\cite{zhang2020multispectral}, LLVIP~\cite{jia2021llvip}, M3FD~\cite{liu2022target}, and DroneVehicle~\cite{9759286}. Sample RGB–IR pairs from these five datasets, as shown in Fig.~\ref{example}. These datasets encompass a diverse range of scenarios, including urban traffic, pedestrian detection, and aerial surveillance, providing a comprehensive benchmark for assessing the performance of our approach.

\vspace{-4pt}

\begin{table*}[htbp]
    \centering
    \caption{Comparisons of different methods on the VEDAI dataset. The best result is shown in \textbf{bold} and the second best is shown with \underline{underline}.}
    \resizebox{\textwidth}{!}{%
    \begin{tabular}{l|cccccccc|c}
        \toprule
        Method (Journal, Year) & Car & Pickup & Camping Car & Truck & Other & Tractor & Boat & Van & $\text{mAP}_{50}$(\%)$\uparrow$ \\
        \midrule
        DMM (\textcolor{red}{TGRS 2025})\cite{zhou2025dmm} & 84.2 & 78.8 & 79.0 & 65.7 & 56.2 & 72.3 & \underline{72.3} & 72.5 & 75.0 \\
        C2DFF-Net (\textcolor{red}{TGRS 2025})\cite{11180153} & \underline{91.8} & \underline{85.9} & \textbf{87.2} & 74.4 & 58.7 & 78.6 & 62.2 & \textbf{91.3} & 79.8 \\
        MRN\_lite (\textcolor{red}{TGRS 2025})\cite{11300300} & \textbf{92.2} & \textbf{87.7} & \underline{83.9} & 73.9 & \underline{68.1} & \underline{80.4} & 54.9 & 72.1 & 76.7 \\
        LGA-YOLO (\textcolor{red}{JSTARS 2025})\cite{zhang2025lga} & 89.4 & \underline{85.9} & 74.1 & \underline{83.4} & 59.9 & \textbf{86.0} & 62.4 & \underline{83.2} & \underline{80.3} \\

        SuperYOLO (\textcolor{red}{TGRS 2023})\cite{10075555} & 91.1 & 85.7 & 79.3 & 70.2 & 57.3 & \underline{80.4} & 60.2 & 76.5 & 75.1 \\
        Ours & 91.5 & 81.0 & 82.0 & \textbf{91.9} & \textbf{78.2} & 80.0 & \textbf{91.9} & 81.3 & \textbf{84.7} \\
        \bottomrule
    \end{tabular}
    }
    \label{tbl:vedai_map50}
    \vspace{-10pt}
\end{table*}

\subsection{Implementation Details}
The model is trained on a single NVIDIA RTX A100 GPU with 80GB of memory. In addition, we conduct inference tests on the NVIDIA RTX 4090 GPU and the Raspberry Pi 5 and compare the frames-per-second (FPS) performance across these three devices.
We implement our algorithm using PyTorch, and adopt the SGD optimizer~\cite{sutskever2013importance} with a momentum of 0.937 and a weight decay of 0.0005. The learning rate is set to 0.01, the batch size is 8, and the model is trained for 300 epochs. Since richer, more comprehensive object annotations are available for the infrared modality, we use the ground truth of IR images as the training labels.

\subsection{Accuracy Metrics}
$\text{mAP}_{50}$ is adopted as the accuracy metric to evaluate the performance of different methods. It is calculated based on precision ($P$) and recall ($R$), which are defined as:
\begin{equation}
P = \frac{TP}{TP + FP}, \quad R = \frac{TP}{TP + FN}
\end{equation}
where $TP$, $FP$, and $FN$ represent the number of true positives, false positives, and false negatives, respectively. A detection is considered a $TP$ if its Intersection over Union (IoU) with the ground truth exceeds a threshold of 0.5. 

The average precision (AP) for a single category is the area under the precision-recall curve:
\begin{equation}
\text{AP}_{50} = \int_{0}^{1} P(R) dR
\end{equation}

Finally, $\text{mAP}_{50}$ is defined as the mean of AP values across all $N$ categories:
\begin{equation}
\text{mAP}_{50} = \frac{1}{N} \sum_{i=1}^{N} \text{AP}_{50, i}
\end{equation}

Additionally, FPS and the number of parameters are used to assess real-time performance and computational cost.

\subsection{Results Comparisons}
\subsubsection{Superiority in Accuracy-Efficiency Trade-off}
\label{sec:cpp}

Tables~\ref{tbl:vedai_map50} and~\ref{tbl:vedai_param_gain} demonstrate that our method achieves a superior trade-off between detection accuracy and model compactness.

\begin{figure*}[htbp]
	\centering
	%\color{red}
	\includegraphics[width=1\textwidth]{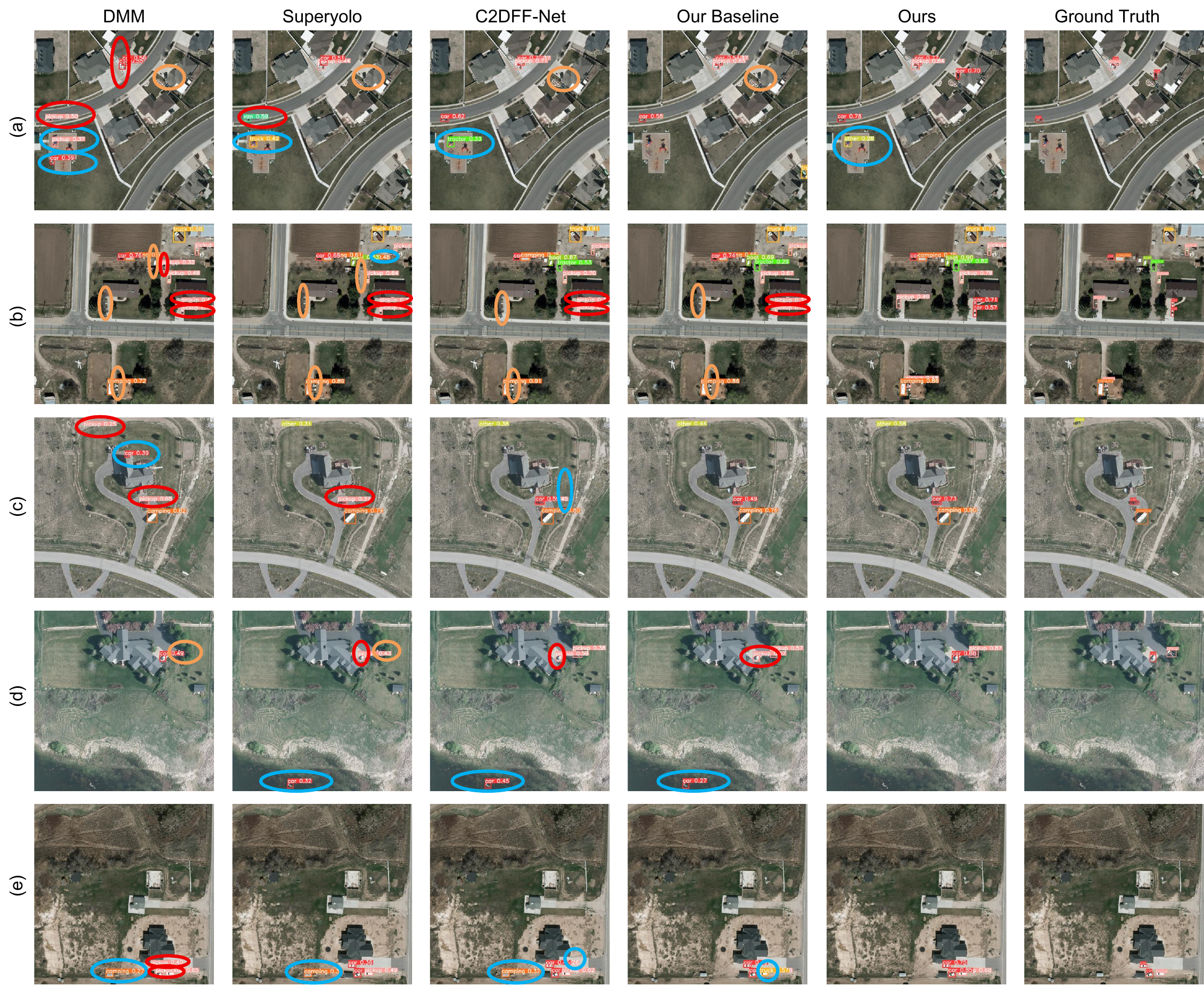}
	\centering
	\caption{Visual comparison of detection results produced by our and competing approaches on the VEDAI dataset under various challenging scenarios. Subfigures (a) and (d) illustrate detection performance in the presence of tree occlusions. Subfigure (b) presents results in an extremely dense scene containing numerous objects of different scales and categories. Subfigures (c) and (e) demonstrate cases where background objects exhibit high visual similarity to the objects. Red circles denote misclassified objects (incorrect category prediction), blue circles indicate false positives (detections of non-existent objects), and yellow circles represent missed detections of ground-truth objects.}
	\label{tab2Comparisons}
    \vspace{-6pt}
\end{figure*}

To further validate the superiority of our approach, we provide a visual comparison of detection results in Fig.~\ref{tab2Comparisons}. While existing mainstream methods frequently suffer from various detection inaccuracies in challenging scenarios, our approach effectively handles tree occlusions and extremely dense scenes, whereas competing models often miss objects or confuse background textures with objects. These qualitative observations are consistent with our quantitative findings, demonstrating that our method not only reduces the parameter count but also learns more robust, semantically consistent feature representations in complex remote sensing environments.

% As summarized in Table~\ref{tbl:vedai_param_gain}, our method achieves a new state-of-the-art on the VEDAI dataset, simultaneously delivering the highest detection accuracy and the most compact model size compared to existing mainstream methods.

\begin{table*}[htbp]
    \centering
    \caption{Comparison of detection accuracy and model size on the VEDAI dataset. The best result is shown in \textbf{bold} and the second best is shown with \underline{underline}.}
    \resizebox{\textwidth}{!}{%
    \begin{tabular}{l|c|c|c|c}
        \toprule
        Method & $\text{mAP}_{50}$(\%)$\uparrow$ & $\Delta\text{mAP}_{50}$ (Method - Ours)$(\%)$& Parameters (M)$\downarrow$ & $\Delta\text{Param}$ (Method - Ours)$(\%)$ \\
        \midrule
        DMM (\textcolor{red}{TGRS 2025})\cite{zhou2025dmm} & 75.00 & -9.70 & 87.97 & +83.53 \\
        C2DFF-Net (\textcolor{red}{TGRS 2025})\cite{11180153} & 79.80 & -4.90 & 6.58 & +2.14 \\
        PG-DRFNet (\textcolor{red}{TCSVT 2025})\cite{wang2025position} & \underline{84.06} & -0.64 & 26.60 & +22.16 \\
        MRN\_lite (\textcolor{red}{TGRS 2025})\cite{11300300} & 76.70 & -8.00 & 6.87 & +2.43 \\
        DETR (\textcolor{red}{JSTARS 2024})\cite{zhu2024cross} & 77.30 & -7.40 & 72.36 & +67.92 \\
        LGA-YOLO (\textcolor{red}{JSTARS 2025})\cite{zhang2025lga} & 80.30 & -4.40 & 10.83 & +6.39 \\
        DKDNet (\textcolor{red}{TGRS 2025})\cite{yue2025diffusion} & 77.90 & -6.80 & 19.20 & +14.76 \\
        CrossModalNet (\textcolor{red}{ESA 2025})\cite{li2025crossmodalnet} & 79.30 & -5.40 & 92.80 & +88.36 \\
        SuperYOLO (\textcolor{red}{TGRS 2023})\cite{10075555} & 75.10 & -9.60& \underline{5.10} & +0.66 \\
        Ours & \textbf{84.70} & 0.00 & \textbf{4.44} & 0.00 \\
        \bottomrule
    \end{tabular}
    }
    \label{tbl:vedai_param_gain}
    \vspace{-10pt}
\end{table*}

\subsubsection{Generalization Across Multiple Benchmarks}
\label{sec:cp}
\begin{figure*}[htbp]
	\centering
	\includegraphics[width=1\textwidth]{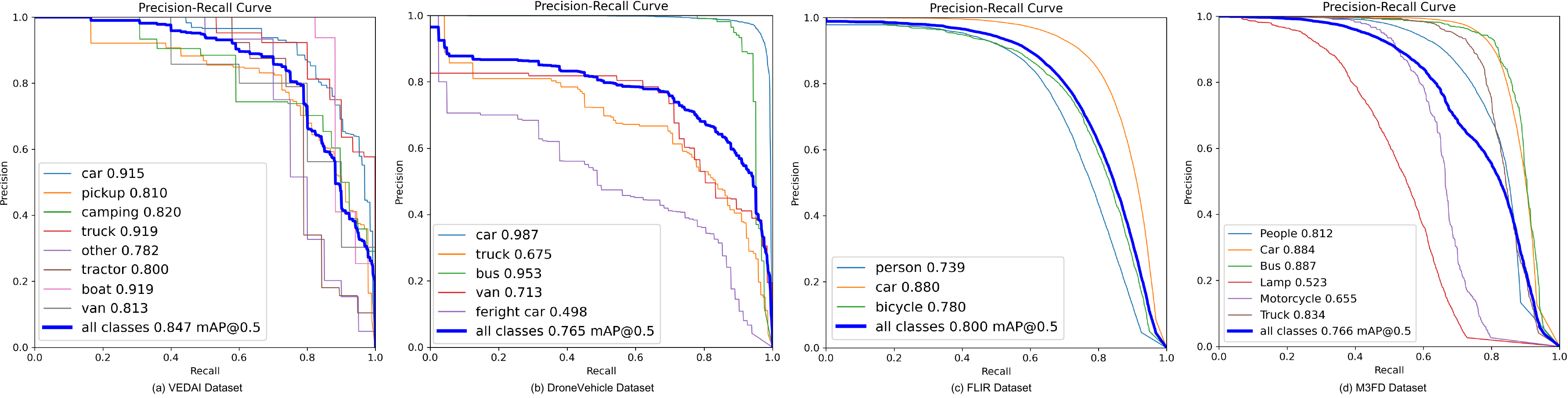}
	\centering
	\caption{Precision–Recall curves of our method on the VEDAI, FLIR, DroneVehicle, and M3FD datasets, showing the detection performance for each class. The LLVIP dataset contains only the person class, so its PR curve is not shown.}
	\label{PRCURVE}
    \vspace{-6pt}
\end{figure*}

To evaluate the robustness of our proposed method, we conduct extensive evaluations across 5 challenging benchmarks. As summarized in Table~\ref{tbl:overall_comparison} and Fig.~\ref{PRCURVE}.
\begin{table*}[htbp]
    \centering
    \caption{Comparisons of different methods on four datasets. The best result for each dataset is shown in \textbf{bold}, and the second-best result is shown in \underline{underline}. The performance gap between our method and the best result is shown in blue.}
    \resizebox{\textwidth}{!}{%
    \begin{tabular}{c|l|c|c|l|c}
        \toprule
        % 双栏表头
        Dataset & Method (Journal, Year) & $\text{mAP}_{50}$(\%)$\uparrow$ & Dataset & Method (Journal, Year) & $\text{mAP}_{50}$(\%)$\uparrow$ \\
        \midrule
        % 左栏：FLIR 数据集
        \multirow{12}{*}{FLIR} 
        & EI2Det (\textcolor{red}{TCSVT 2025})\cite{hu2025ei} & \textbf{80.2} & 
        % 右栏：LLVIP 数据集
        \multirow{12}{*}{LLVIP} 
        & EI2Det (\textcolor{red}{TCSVT 2025})\cite{hu2025ei} & \textbf{98.0} \\
        
        & MMI-Det (\textcolor{red}{TCSVT 2024})\cite{10570450} & 79.8 & 
        & DHANet (\textcolor{red}{TGRS 2025})\cite{wu2025dhanet} & \underline{97.7} \\
        
        & ICAFusion (\textcolor{red}{PR 2024})\cite{shen2024icafusion} & 77.5 & 
        & PSFusion (\textcolor{red}{IF 2023})\cite{tang2023rethinking} & 96.6 \\
        
        & PSFusion (\textcolor{red}{IF 2023})\cite{tang2023rethinking} & 75.8 & 
        & YOLO-Adaptor (\textcolor{red}{TIV 2024})\cite{fu2024yolo} & 96.5 \\
        
        & CDDFuse (\textcolor{red}{CVPR 2023})\cite{zhao2023cddfuse} & 75.5 & 
        & PIAFusion (\textcolor{red}{IF 2022})\cite{tang2022piafusion} & 96.1 \\
        
        & DATFuse (\textcolor{red}{TCSVT 2023})\cite{tang2023datfuse} & 75.4 & 
        & CDDFuse (\textcolor{red}{CVPR 2023})\cite{zhao2023cddfuse} & 95.7 \\
        
        & PIAFusion (\textcolor{red}{IF 2022})\cite{tang2022piafusion} & 75.3 & 
        & IFCNN (\textcolor{red}{IF 2020})\cite{zhang2020ifcnn} & 95.5 \\
        
        & IFCNN (\textcolor{red}{IF 2020})\cite{zhang2020ifcnn} & 74.9 & 
        & YOLO Fusion (\textcolor{red}{PR 2022})\cite{qingyun2022cross} & 95.4 \\
        
        & DHANet (\textcolor{red}{TGRS 2025})\cite{wu2025dhanet} & 74.3 & 
        & MoE-Fusion (\textcolor{red}{ICCV 2023})\cite{cao2023multi} & 91.0 \\
        
        & GAFF (\textcolor{red}{WACV 2021})\cite{zhang2021guided} & 72.9 & 
        & DIVFusion (\textcolor{red}{IF 2023})\cite{tang2023divfusion} & 89.8 \\
        
        & YOLO Fusion (\textcolor{red}{PR 2022})\cite{qingyun2022cross} & 71.7 & 
        & DM-Fusion (\textcolor{red}{TNNLS 2024})\cite{xu2023dm} & 88.1 \\
        
        & Ours & \underline{80.0} {\footnotesize\textcolor{blue}{(-0.2)}} & 
        & Ours & 97.5 {\footnotesize\textcolor{blue}{(-0.5)}} \\
        
        \midrule
        % 左栏：M3FD 数据集
        \multirow{7}{*}{M3FD} 
        & MMI-Det (\textcolor{red}{TCSVT 2024})\cite{10570450} & \textbf{76.6} & 
        % 右栏：DroneVehicle 数据集
        \multirow{7}{*}{\begin{tabular}{c} Drone \\ Vehicle \end{tabular}} 
        & CIAN (\textcolor{red}{IF 2019})\cite{zhang2019cross} & 70.8 \\
        
        & ICAFusion (\textcolor{red}{PR 2024})\cite{shen2024icafusion} & \underline{71.9} & 
        & AR-CNN (\textcolor{red}{TNNLS 2021})\cite{zhang2021weakly} & 71.6 \\
        
        & PIAFusion (\textcolor{red}{IF 2022})\cite{tang2022piafusion} & 69.9 & 
        & TSFADet (\textcolor{red}{ECCV 2022})\cite{yuan2022translation} & 73.1 \\
        
        & PSFusion (\textcolor{red}{IF 2023})\cite{tang2023rethinking} & 69.7 & 
        & $C\textsuperscript{2}Former$ (\textcolor{red}{TGRS 2024})\cite{yuan2024c2former} & \underline{74.2} \\
        
        & CDDFuse (\textcolor{red}{CVPR 2023})\cite{zhao2023cddfuse} & 69.5 & 
        & $S\textsuperscript{2}ANET$ (\textcolor{red}{TGRS 2021})\cite{han2021align} & 71.5 \\
        
        & IFCNN (\textcolor{red}{IF 2020})\cite{zhang2020ifcnn} & 69.0 & 
        & MBNet (\textcolor{red}{ECCV 2020})\cite{zhou2020improving} & 71.9 \\
        
        & Ours & \textbf{76.6} & 
        & Ours & \textbf{76.5} {\footnotesize\textcolor{blue}{(+2.3)}} \\
        
        \bottomrule
    \end{tabular}
    }
    \label{tbl:overall_comparison}
    \vspace{-6pt}
\end{table*}

\subsubsection{Superiority over pruning compression techniques}
\label{sec:c}
Table~\ref {tbl:compression} shows that our Low-Rank SS2D method achieves a better balance between accuracy and efficiency compared with existing pruning compression approaches.
\begin{table}[H]  % 改为单栏表格环境table（table*是双栏跨栏环境）
    \centering
    \caption{Comparison of our LowRank-SS2D method with mainstream existing lightweighting methods on the VEDAI dataset. (I: SS2D baseline; II: SS2D + pruning; III: SS2D + Our LowRank; FPS measured on Raspberry Pi 5, FP32; blue parentheses indicate percentage change relative to I).}
    \label{tbl:compression}
    % 删除model列，列数从6列改为5列，适配单栏宽度
    \resizebox{\columnwidth}{!}{%
    \begin{tabular}{c c c c c}
        \toprule
        No. & SS2D replaced & Params (MB)$\downarrow$ & $\text{mAP}_{50}$(\%)$\uparrow$ & FPS$\uparrow$ \\
        \midrule
        I   & 0 & 17.1 & \textbf{81.5} & 0.4 \\
        II  & 0 & \underline{9.0} {\scriptsize(\textcolor{blue}{-47.4\%})} & 70.2 {\scriptsize(\textcolor{blue}{-13.9\%})} & \underline{1.0} {\scriptsize(\textcolor{blue}{+150.0\%})} \\
        III & 6 & \textbf{8.5} {\scriptsize(\textcolor{blue}{-50.3\%})} & \underline{75.5} {\scriptsize(\textcolor{blue}{-7.4\%})} & \textbf{1.2} {\scriptsize(\textcolor{blue}{+200.0\%})} \\
        \bottomrule
    \end{tabular}
        }
    \vspace{-8pt}
\end{table}
% \begin{table*}[htbp]
%     \centering
%     \caption{Comparison of our LowRank-SS2D method with mainstream existing model compression and lightweighting methods on the VEDAI dataset (FPS on Raspberry Pi 5, FP32; blue text in parentheses denotes percentage change relative to Baseline (I)).}
%     \label{tbl:compression}
%     \begin{tabular}{c l c c c c}
%         \toprule
%         No. & Model & SS2D replaced & Params (MB)$\downarrow$ & $\text{mAP}_{50}$(\%)$\uparrow$ & FPS$\uparrow$ \\
%         \midrule
%         I   & Original SS2D (Baseline) & 0 & 17.1 & \textbf{81.5} & 0.4 \\
%         II  & SS2D + Pruning & 0 & \underline{9.0} {\scriptsize(\textcolor{blue}{-47.4\%})} & 70.2 {\scriptsize(\textcolor{blue}{-13.9\%})} & \underline{1.0} {\scriptsize(\textcolor{blue}{+150.0\%})} \\
%         III & SS2D + Our LowRank & 6 & \textbf{8.5} {\scriptsize(\textcolor{blue}{-50.3\%})} & \underline{75.5} {\scriptsize(\textcolor{blue}{-7.4\%})} & \textbf{1.2} {\scriptsize(\textcolor{blue}{+200.0\%})} \\
%         \bottomrule
%     \end{tabular}
%     \vspace{-8pt}
% \end{table*}

\subsubsection{Cross-Platform Efficiency Comparison}
\label{sec:cross_platform}
Table~\ref{tbl:inference_speed_devices} demonstrates the superior efficiency of our proposed method across diverse hardware architectures compared to the baseline. While our approach consistently outperforms the baseline on high-end GPUs, its most significant advantage is observed in resource-constrained environments. On the Raspberry Pi 5, our method achieves a 5.5$\times$ speedup in FPS.
\begin{table}[H]
    \centering
    \caption{Inference speed on different devices.}
    \label{tbl:inference_speed_devices}

    \resizebox{\columnwidth}{!}{%
    \begin{tabular}{c l l c c}
        \toprule
        No. & Method & Device & FPS$\uparrow$ & Inference time (ms)$\downarrow$ \\
        \midrule
        \multirow{2}{*}{I}   & Baseline & \multirow{2}{*}{GPU A100}        & \underline{41.88} & \underline{23.88} \\
                            & Ours     &                                  & \textbf{43.59}     & \textbf{22.94} \\
        \midrule
        \multirow{2}{*}{II}  & Baseline & \multirow{2}{*}{GPU 4090}        & \underline{19.60} & \underline{51.03} \\
                            & Ours     &                                  & \textbf{29.00}     & \textbf{34.49} \\
        \midrule
        \multirow{2}{*}{III} & Baseline & \multirow{2}{*}{Raspberry Pi 5} & \underline{0.42}  & \underline{2367.67} \\
                            & Ours     &                                  & \textbf{2.30}      & \textbf{434.78} \\
        \bottomrule
    \end{tabular}
    }
\end{table}

\subsection{Ablation Study}
\label{sec: Ablation Study}
\subsubsection{Relationship Between Low-Rank Decomposition Ratio and Inference Efficiency}

We analyze the sensitivity of our method to the low-rank decomposition ratio, a critical hyperparameter that balances detection accuracy and edge-device efficiency. As Table~\ref{tbl:rank_ratio_tradeoff} shows, performance varies predictably with the rank ratio. A high ratio (0.65, Row No. I) achieves the highest accuracy (80.20\% $\text{mAP}_{50}$) but incurs a substantial computational cost on devices such as the Raspberry Pi 5 (0.64 FPS). Reducing the ratio markedly increases inference speed, with only minor degradation in accuracy. Under strict real-time constraints (Row No. IV), a ratio of 0.50 nearly doubles the FPS (+90.6\%) while only reducing $\text{mAP}_{50}$ by 4.67\%. These results demonstrate the method’s robustness to rank adjustments and its flexibility for tailoring performance to edge-resource limitations.
\begin{table}[!t]
    \centering
    \caption{Effect of low-rank decomposition rank ratio on detection accuracy and Raspberry Pi 5 inference speed.}
    \label{tbl:rank_ratio_tradeoff}
    \resizebox{\columnwidth}{!}{%
    \begin{tabular}{c c c c}
        \toprule
        No. & Rank ratio & $\text{mAP}_{50}$(\%)$\uparrow$ & FPS (Raspberry Pi 5)$\uparrow$ \\
        \midrule
        I   & 0.65 & 80.20& 0.64  \\
        II  & 0.60 & 79.60 & 0.98 {\scriptsize(\textcolor{blue}{+53.1\%})} \\
        III & 0.55 & 76.71 & 1.13 {\scriptsize(\textcolor{blue}{+76.6\%})} \\
        IV  & 0.50 & 75.53 & 1.22 {\scriptsize(\textcolor{blue}{+90.6\%})} \\
        \bottomrule
    \end{tabular}
    }
    \vspace{-4pt}
\end{table}

% \subsubsection{\textbf{XXX}} 
% % In Table \ref{xxx}, the model 

% To meet the real-time requirement on resource-constrained devices, we deliberately adopt a more aggressive low-rank configuration and rely on distillation and task-specific fine-tuning to restore performance. Specifically, we use the full-rank SS2D model as the teacher and distill its knowledge into the low-rank student. As reported in \cref{tbl:ablation_pi5}, the initial Low rank-SS2D increases the inference speed from 0.4 to 1.2 FPS and reduces the model size from 17.1 to 8.5 MB, but its $\text{mAP}_{50}$ drops from 81.5\% to 75.5\%; after full-rank-to-low-rank distillation followed by fine-tuning, the resulting compact model remains at 4.3 MB while achieving 2.3 FPS and improving $\text{mAP}_{50}$ to 84.7\%.
\subsubsection{Effectiveness of Structure-Aware Distillation for Low-Rank SS2D}
To meet the real-time requirement on resource-constrained devices, we deliberately adopt a more aggressive low-rank configuration. We rely on distillation and fine-tuning to restore performance.

Our ablation study, as shown in Table~\ref{tbl:ablation_pi5}, underscores the critical role of the proposed structure-aware distillation in mitigating performance degradation and enhancing inference efficiency. Without the tailored distillation strategy (Row No. II), the initial low-rank decomposition of the SS2D layers reduces the parameter count from 17.1 MB to 8.5 MB. This decomposition, however, results in a significant accuracy drop of 6.0\% $\text{mAP}_{50}$ compared to the baseline (Row No. I). This indicates that low-rank decomposition alone cannot preserve high-fidelity feature representations. Upon integrating structure-aware distillation (Row No. III and No. IV), we observe substantial performance gains. Notably, the distilled student model not only achieves a significant speedup in inference. It increases from 0.4 FPS to 2.3 FPS on Raspberry Pi 5. When combined with fine-tuning, this model also surpasses the baseline accuracy by 3.2\% $\text{mAP}_{50}$ (Row No. IV vs. Row No. I). This suggests that the distillation process effectively transfers the complex spatial-spectral correlations from the original SS2D layers into the compact low-rank counterparts.
\begin{table}[!t]
    \centering
    \caption{Ablation study of model compression and acceleration on VEDAI dataset (FPS on Raspberry Pi 5, FP32). I: Original SS2D (Baseline); II: LowRank-SS2D without distillation; III: LowRank-SS2D with distillation; IV: LowRank-SS2D with distillation and fine-tuning. The best result is shown in \textbf{bold} and the second best is shown with \underline{underline}.}
    \resizebox{\columnwidth}{!}{%
    \begin{tabular}{c c c c c}
        \toprule
        No. & SS2D replaced & Params (MB)$\downarrow$ & $\text{mAP}_{50}$(\%)$\uparrow$ & FPS$\uparrow$ \\
        \midrule
        I   & 0 & 17.1 & \underline{81.5} & 0.4 \\
        II   & 6 & \underline{8.5} & 75.5 & \underline{1.2} \\
        III  & 6 & \textbf{4.3} & 80.1 & \textbf{2.3} \\
        IV  & 6 & \textbf{4.3} & \textbf{84.7} & \textbf{2.3} \\
        \bottomrule
    \end{tabular}
    }
    \label{tbl:ablation_pi5}
    \vspace{-8pt}
\end{table}

\begin{figure}[!t]
	\centering
	\includegraphics[width=0.48\textwidth]{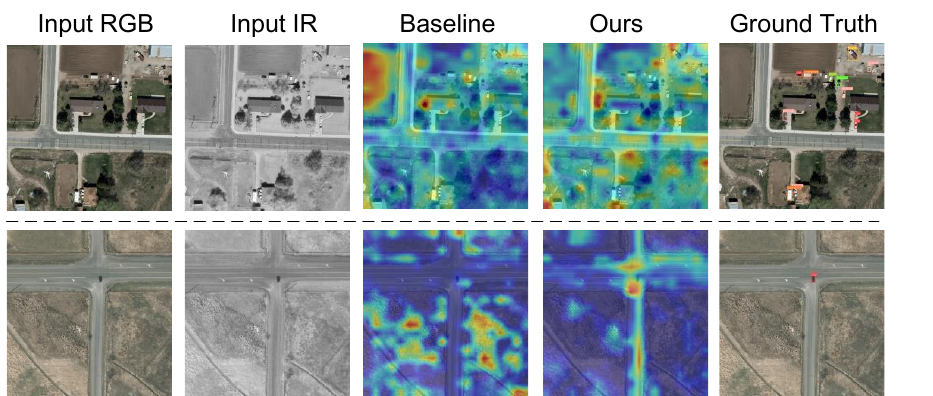}
	\centering
	\caption{Comparison of Grad-CAM heatmaps generated from the SS2D layer for the baseline and our method.}
	\label{attention}
    \vspace{-6pt}
\end{figure}

Compared with the original baseline, the model optimized with our structure-aware distillation exhibits more concentrated, semantically consistent activation patterns. This indicates that our method enables the low-rank model to better focus on discriminative object features, explaining the observed quantitative improvements in detection accuracy. As shown in Fig.~\ref{attention}.

\section{Conclusion and Future Work}
\label{sec:Conclusion}

%%%%%%%%%qianqian todo
In this paper, we present an efficient multispectral object detection framework based on 2D Selective Structured State Space (SS2D) models, specifically optimized for edge deployment. 
First, we propose the Low-Rank SS2D backbone, which reconstructs the vanilla full-rank SS2D via matrix factorization. This design achieves significant model compression while preserving linear complexity and global receptive fields. 
Second, to alleviate the representation degradation incurred by low-rank approximation, we introduce a Structure-Aware Distillation strategy. By aligning the singular components and hidden-state dynamics between the full-rank teacher and the low-rank student, our method recovers critical fine-grained structural information for detection. 
Extensive experiments across five benchmarks and diverse hardware platforms, ranging from high-end GPUs to resource-constrained edge devices, demonstrate that our approach achieves competitive accuracy with significantly enhanced inference efficiency. 
Future work will explore adaptive low-rank configurations to further push the Pareto frontier of efficiency and precision on the edge.

\ifCLASSOPTIONcaptionsoff
\newpage
\fi
{
	\bibliographystyle{IEEEtran}
	\bibliography{reference}
}

\vspace{-2em}

\vspace{-1em}
\begin{IEEEbiography}[{\includegraphics[width=0.9in,clip,keepaspectratio]{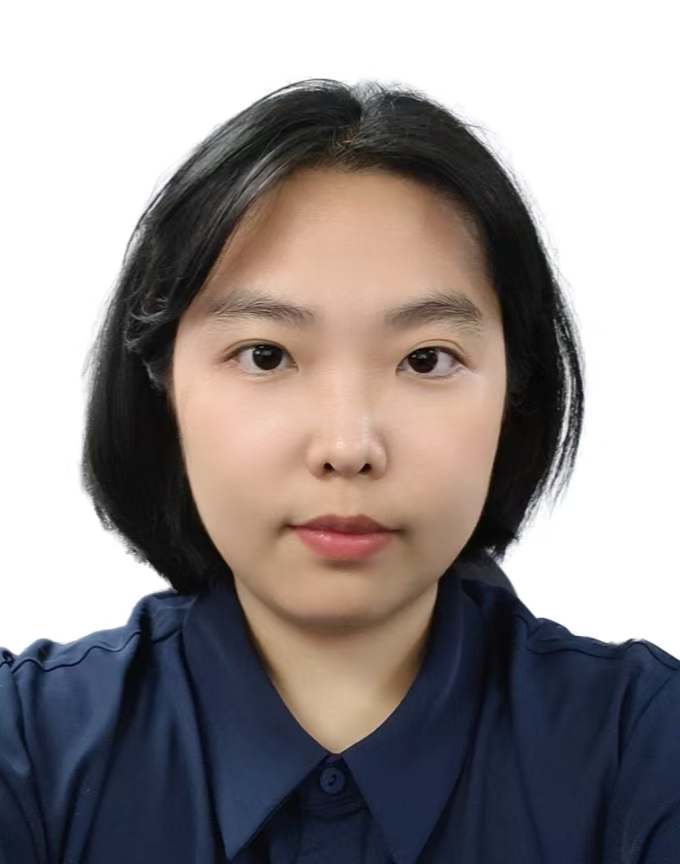}}]{Qianqian Zhang}
	is currently a Ph.D. candidate specializing in Computer Application Technology at the University of Chinese Academy of Sciences (UCAS), Beijing, China. Supported by the China Scholarship Council, she is also conducting joint training and collaboration with Queen Mary University of London, London, United Kingdom.
	
	Her research interests focus on large visual models, multimodal fusion, object detection, model lightweighting, efficient deployment, and video compression.
\end{IEEEbiography}

\vspace{-1em}
\begin{IEEEbiography}[{\includegraphics[width=0.9in,clip,keepaspectratio]{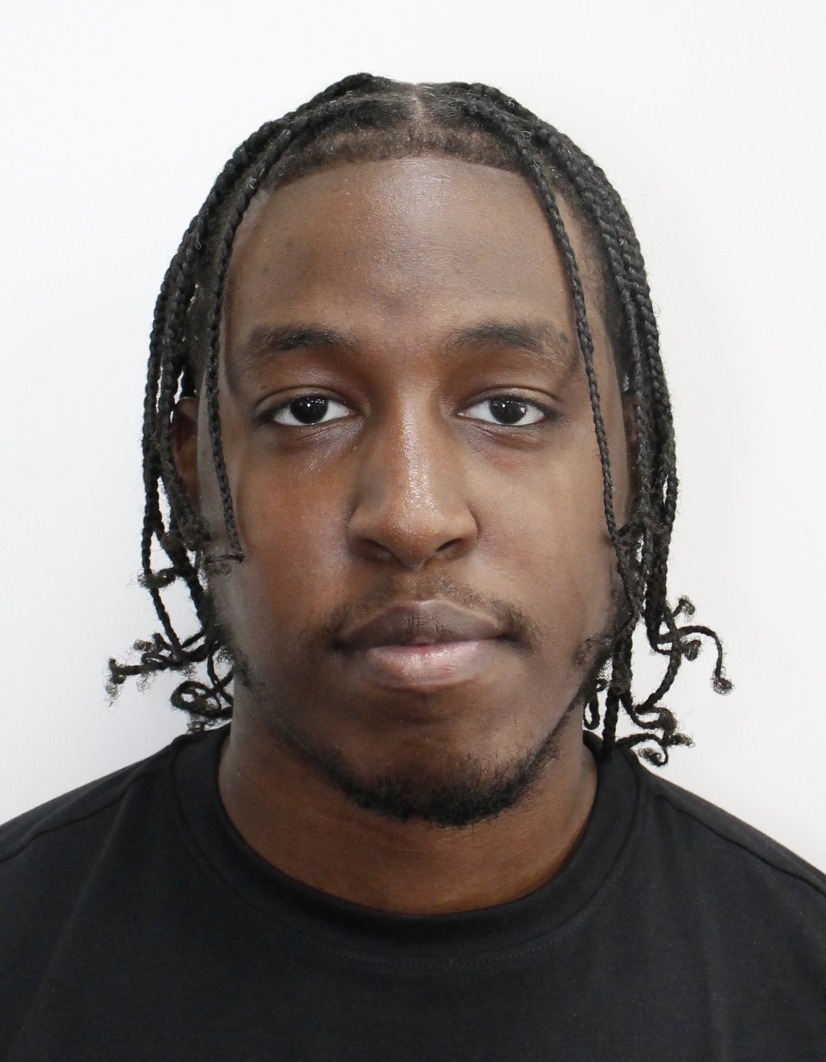}}]{Leon Tabaro}
	is currently a Ph.D. candidate at the School of Electronic Engineering and Computer Science, Queen Mary University of London, London, UK. His research interests lie at the intersection of geometric deep learning, machine learning systems and optimization, with a focus on general purpose deep sequence models with long-range memory, efficient training and inference, and structured sparsity for compact deep learning models.
    
\end{IEEEbiography}

\vspace{-1em}
\begin{IEEEbiography}[{\includegraphics[width=0.9in,clip,keepaspectratio]{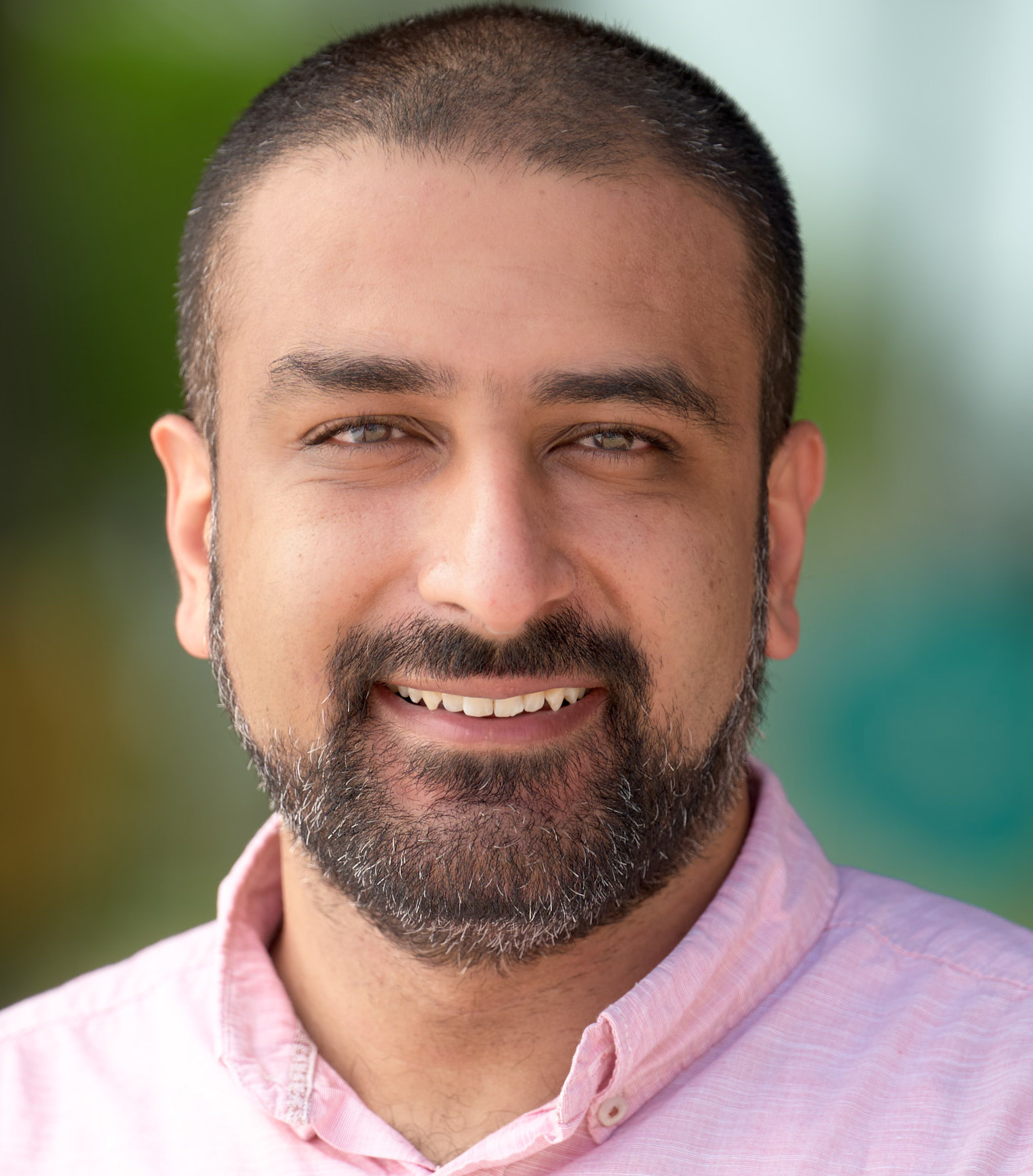}}]{Ahmed M. Abdelmoniem}
	(Senior Member, IEEE) is an Associate Professor at the School of Electronic Engineering and Computer Science, Queen Mary University of London, UK, and leads SAYED Systems Group. He held the positions of Research Scientist at KAUST, Saudi Arabia, and Senior Researcher at Huawei's Future Networks Lab (FNTL), Hong Kong. He is the principal investigator and co-investigator on several national and international research projects, funded mainly by grants totaling over USD\$1.8 million. He received his PhD in Computer Science and Engineering from the Hong Kong University of Science and Technology, Hong Kong, in 2017. He was awarded the prestigious Hong Kong PhD Fellowship from the RGC of Hong Kong in 2013 to pursue his PhD at HKUST, HK. He has published numerous (more than 135) papers in top venues and journals in distributed systems, computer networking, and machine learning. His current research interests are optimizing systems supporting distributed machine learning, federated learning, and cloud/data-center networking, emphasizing performance, practicality, and scalability.
\end{IEEEbiography}

\vspace{-1em}
\begin{IEEEbiography}[{\includegraphics[width=0.9in,clip,keepaspectratio]{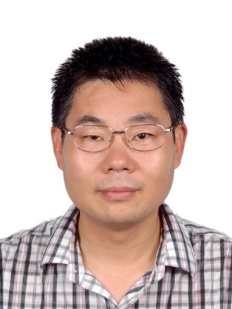}}]{Junshe An}
	received the Bachelor of Science degree from Beihang University, Beijing, China, in 1992, the Master of Science degree from the University of Science and Technology Beijing, Beijing, China, in 1995, and the Doctor of Philosophy degree from Northwestern Polytechnical University, Xi'an, China, in 2004. He is currently a Researcher with the National Space Science Center, Chinese Academy of Sciences, Beijing, China.
	
	His research interests focus on space-integrated electronic technologies, including aerospace computer hardware and software, system architecture, and intelligent information processing.
\end{IEEEbiography}

\end{document}